\documentclass[9pt,technote]{IEEEtran}
\usepackage{amssymb}
\usepackage{amsmath}
\usepackage{amsfonts}
\usepackage{graphicx}%
\providecommand{\U}[1]{\protect\rule{.1in}{.1in}}
\providecommand{\U}[1]{\protect\rule{.1in}{.1in}}
\newtheorem{theorem}{Theorem}

\newtheorem{definition}[theorem]{Definition}

\begin{document}

\title{Fuzzy quantification for linguistic data analysis and data mining.}
\author{F. D\'{\i}az-Hermida, Juan. C. Vidal\ \\\{felix.diaz,juan.vidal\}@usc.es\\Centro Singular de Investigaci\'{o}n en Tecnolox\'{\i}as da Informaci\'{o}n
(CiTIUS), Universidade de Santiago de Compostela, Spain.}
\maketitle

\begin{abstract}
Fuzzy quantification is a subtopic of fuzzy logic which deals with the
modelling of the quantified expressions we can find in natural language. Fuzzy
quantifiers have been successfully applied in several fields like fuzzy,
control, fuzzy databases, information retrieval, natural language generation,
etc. Their ability to model and evaluate linguistic expressions in a
mathematical way, makes fuzzy quantifiers very powerful for data analytics and
data mining applications. In this paper we will give a general overview of the
main applications of fuzzy quantifiers in this field as well as some ideas to
use them in new application contexts.

\end{abstract}

\textbf{Keywords:} fuzzy quantification, theory of generalized quantifiers,
data analysis, data mining

\textit{Published in the III Data Science \& Engineering Consortium Meeting}

\section{Introduction}

\begin{figure}
[ptb]
\begin{center}
\includegraphics[
height=4.1816cm,
width=5.3225cm
]%
{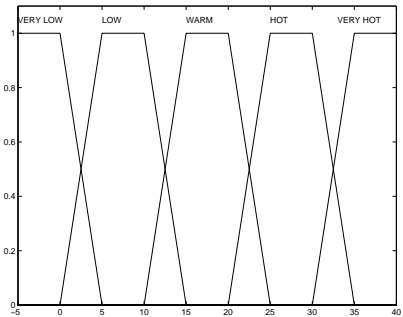}%
\caption{Linguistic labels associated to the linguistic variable
`temperature'.}%
\label{FiguTemperatureLabels}%
\end{center}
\end{figure}

Fuzzy logic \cite{Zadeh65} is a subfield of artificial intelligence that deals
with the management of vague and imprecise expressions. In fuzzy logic
systems, the classical logic based on binary truth values is generalized to
fuzzy truth values defined on the interval $\left[  0,1\right]  $. Classical
logical operators are substituted with families of fuzzy operators that
generalize them to fuzzy truth values, and sets are generalized to `fuzzy
sets', where belongingness cease to be a `classic or crisp' concept to become
a fuzzy concept. In this way, fuzzy sets accept partial fulfillment of their
elements. For example, the fuzzy set of `tall people' can include people that
is tall only to a partial degree. A crucial concept in fuzzy logic is the
concept of linguistic variable \cite{Zadeh75}, which allows to divide the
range of variation of a variable (e.g., `temperature') by means of fuzzy
linguistic labels (e.g. `very low', `low', `warm', `hot', `very hot', see
Figure \ref{FiguTemperatureLabels}).%

%TCIMACRO{\FRAME{ftbpFU}{5.3225cm}{4.1816cm}{0pt}{\Qcb{Linguistic labels
%associated to the linguistic variable `temperature'.}}%
%{\Qlb{FiguTemperatureLabels}}{temperature_labels.eps}%
%width 5.3225cm;  height 4.1816cm;  depth 0pt;  original-width 6.2914in;
%original-height 4.9248in;  cropleft "0";  croptop "1";  cropright "1";
%filename 'Imagenes/temperature_labels.eps';file-properties "XNPEU";}} }%
%BeginExpansion
%EndExpansion

This paper deals with the specific application of fuzzy quantifiers to data
analytics and data mining. Fuzzy quantifiers were introduced by Zadeh
\cite{Zadeh83} to model \textit{quantified linguistic expressions.} In his
approach, Zadeh distinguished two types of linguistic quantifiers:
\textit{quantifiers of the first kind,\thinspace}used to represent absolute
quantities (defined by using fuzzy numbers on $\mathbb{N}$), and
\textit{quantifiers of the second kind, }used to represent relative quantities
(defined by using fuzzy numbers on $\left[  0,1\right]  $).

In this work, we will follow Gl\"{o}ckner's approximation to fuzzy
quantification \cite{Glockner06Libro}. In his approach, the author generalizes
the concept of \textit{generalized classic quantifier} \cite{Barwise81}
(second order predicates or set relationships) to the fuzzy case; that is, a
\textit{fuzzy quantifier }is a fuzzy relationship between fuzzy sets. After
generalizing the concept of classic quantifier to the fuzzy case, he rewrote
the fuzzy quantification problem as a problem of looking for possible
mechanisms to convert \textit{semi-fuzzy quantifiers} (quantifiers occupying a
middle point between generalized classic quantifiers and fuzzy quantifiers) to
fuzzy quantifiers. The author called these transfomation mechanisms
\textit{Quantifier Fuzzification Mechanism} (\textit{QFMs}). Following the
linguistic \textit{Theory of Generalized Quantifiers (TGQ)} \cite{Barwise81},
this approach is capable of handling most of the quantification phenomena of
natural language. In addition, it also allows the translation of most of the
analysis that has been made in TGQ from a linguistic perspective to the fuzzy
case, facilitating the definition and the test of adequacy properties.
Gl\"{o}ckner has also defined a rigorous axiomatic framework to ensure the
good behavior of QFMs. Models fulfilling this framework are called
\textit{Determiner fuzzification schemes (DFSs) }and they comply with a broad
set of properties that guarantee a good behavior from a linguistic and fuzzy
point of view. See the recent \cite{Sanchez16} or \cite{Glockner06Libro} for a
comparison between Zadeh's and Gl\"{o}ckner's approaches.

The objective of this paper is to present some of the different roles that
fuzzy quantification can play in data analytics and data mining. First, fuzzy
quantification can be used in a `descriptive sense'. In this case, fuzzy
quantifiers are simply used to model some linguistic expression that can be of
utility in a particular domain. We will explicitly show some examples of the
application of fuzzy quantifiers in the temporal domain, to prove the capacity
of fuzzy quantifiers to model `quantified temporal expressions' (e.g.,
\textit{\textquotedblleft the temperature was low for most of the last
minutes\textquotedblright}).

Second, fuzzy quantifiers can be used in a `summarization sense'. In this
case, we are interested in automatically computing a single quantifier or a
set of quantified expressions to summarize a set of data (e.g. to infer that
the quantifier\textit{\ }`most'\textit{ }is the one that better explained the
amount of \textit{`}warm temperatures in June'). Different problems arise in
the summarization of data by means of fuzzy quantifiers, as we will see
throughout the paper.

Finally, fuzzy quantifiers can be used in combination with other techniques in
machine learning problems. We will show some specific examples of the
application of fuzzy quantifiers for learning fuzzy quantified constraint
networks and fuzzy quantified systems of rules. Some of the examples we will
present in this paper have not been theoretically developed yet, and they are
presented just as an idea of the power of fuzzy quantifiers to be combined
with other techniques.

\section{The fuzzy quantification framework}

Most approaches to fuzzy quantification follow the concept of \textit{fuzzy
linguistic quantifier, }which was proposed to represent absolute or
proportional fuzzy quantities. Following Zadeh \cite{Zadeh83},
\textit{quantifiers of the first kind\thinspace}are the adequate mean to
represent absolute quantities (by using fuzzy numbers on $\mathbb{N}$), whilst
\textit{quantifiers of the second kind }are the adequate mean to represent
relative quantities (by using fuzzy numbers on $\left[  0,1\right]  $)

As we mentioned before, in this paper, we will follow the approximation to
fuzzy quantification proposed in \cite{Glockner06Libro}. Let us introduce now
some definitions:

\begin{definition}
A two valued (generalized) quantifier on a base set $E\neq\varnothing$ is a
mapping $Q:\mathcal{P}\left(  E\right)  ^{n}\longrightarrow\mathbf{2}$, where
$n\in\mathbb{N}$ is the arity (number of arguments) of $Q$, $\mathbf{2}%
=\left\{  0,1\right\}  $ denotes the set of crisp truth values, and
$\mathcal{P}\left(  E\right)  $ is the powerset of $E$.
\end{definition}

Here we present two examples of classic quantifiers:%
\begin{align*}
\mathbf{all}\left(  Y_{1},Y_{2}\right)   &  =Y_{1}\subseteq Y_{2}\\
\mathbf{at\_least}60\%\left(  Y_{1},Y_{2}\right)   &  =\left\{
\begin{array}
[c]{cc}%
\frac{\left\vert Y_{1}\cap Y_{2}\right\vert }{\left\vert Y_{1}\right\vert
}\geq0.60 & Y_{1}\neq\varnothing\\
1 & Y_{1}=\varnothing
\end{array}
\right.
\end{align*}

In a fuzzy quantifier, inputs and outputs can be fuzzy. They assign a gradualt
result to each choice of $X_{1},\ldots,X_{n}\in\widetilde{\mathcal{P}}\left(
E\right)  $, where by $\widetilde{\mathcal{P}}\left(  E\right)  $ we denote
the fuzzy powerset of $E$ (i.e., the set of all possible fuzzy sets of $E$).

\begin{definition}
\cite{Glockner06Libro} An n-ary fuzzy quantifier $\widetilde{Q}$ on a base set
$E\neq\varnothing$ is a mapping $\widetilde{Q}:\widetilde{\mathcal{P}}\left(
E\right)  ^{n}\longrightarrow\mathbf{I=}\left[  0,1\right]  $.
\end{definition}

For example, the fuzzy quantifier $\widetilde{\mathbf{all}}:\widetilde
{\mathcal{P}}\left(  E\right)  ^{2}\longrightarrow\mathbf{I}$ could be defined
as:
\[
\widetilde{\mathbf{all}}\left(  X_{1},X_{2}\right)  =\inf\left\{  \max\left(
1-\mu_{X_{1}}\left(  e\right)  ,\mu_{X_{2}}\left(  e\right)  \right)  :e\in
E\right\}
\]
where by $\mu_{X}\left(  e\right)  $ we denote the membership function of
$X\in\widetilde{\mathcal{P}}\left(  E\right)  $. Here, $\inf\left\{
\max\left(  1-\mu_{X_{1}}\left(  e\right)  ,\mu_{X_{2}}\left(  e\right)
\right)  :e\in E\right\}  $ represents the generalization, to the fuzzy world
of the logical set inclusion defined as $\forall e,X_{1}\left(  e\right)
\rightarrow X_{2}\left(  e\right)  =\forall e,\lnot X_{1}\left(  e\right)
\vee X_{2}\left(  e\right)  $. The $\lnot$ operator is converted into the
fuzzy strong negation $\widetilde{\lnot}\left(  x\right)  =1-x$, the $\vee$
operator into the standard $max$ tconorm $\widetilde{\vee}\left(  x_{1}%
,x_{2}\right)  =\max\left(  x_{1},x_{2}\right)  $ and the $\forall$ operator
into the infimum, which for finite base sets coincides with the minimun.

In the fuzzy case other options to define the quantifier $\widetilde
{\mathbf{all}}$ could be considered (e.g. by substituting the negation
operator or the tconorm operator by other reasonable fuzzy alternatives). But
in general, the problem of defining adequate fuzzy quantification models can
be quite difficult (e.g., look for a reasonable expression to evaluate
\textit{`at least sixty percent'}). In \cite{Glockner06Libro} this problem is
faced by introducing the concept of semi-fuzzy quantifiers, that could be
interpreted as a middle ground between classic quantifiers and fuzzy
quantifiers. Semi-fuzzy quantifiers take as input crisp arguments, as classic
quantifiers, but allow the result to range over the truth grade scale
$\mathbf{I}$, as in the case of fuzzy quantifiers.

\begin{definition}
\cite{Glockner06Libro} An n-ary semi-fuzzy quantifier $Q$ on a base set
$E\neq\varnothing$ is a mapping $Q:\mathcal{P}\left(  E\right)  ^{n}%
\longrightarrow\mathbf{I}$.
\end{definition}

\begin{figure}
[ptb]
\begin{center}
\includegraphics[
height=1.9169cm,
width=5.3225cm
]%
{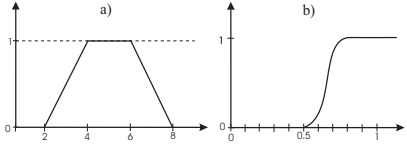}%
\caption{Fuzzy numbers for modelling \textit{`around 5'} and \textit{`at lest
80\%'}.}%
\label{FiguQuantifiers}%
\end{center}
\end{figure}

For each pair of crisp sets $\left(  Y_{1},\ldots,Y_{n}\right)  $, $Q\left(
Y_{1},\ldots,Y_{n}\right)  $ is a gradual value. Two examples of semi-fuzzy
quantifiers are:
\begin{align}
\mathbf{about\_5}\left(  Y_{1},Y_{2}\right)   &  =T_{2,4,6,8}\left(
\left\vert Y_{1}\cap Y_{2}\right\vert \right) \label{about_or_more_80}\\
\mathbf{at\_least}\_\mathbf{about}80\%\left(  Y_{1},Y_{2}\right)   &
=\left\{
\begin{array}
[c]{cc}%
S_{0.5,0.8}\left(  \frac{\left\vert Y_{1}\cap Y_{2}\right\vert }{\left\vert
Y_{1}\right\vert }\right)  & X_{1}\neq\varnothing\\
1 & X_{1}=\varnothing
\end{array}
\right. \nonumber
\end{align}
where $T_{2,4,6,8}\left(  x\right)  $ and $S_{0.5,0.8}\left(  x\right)  $
represent two common fuzzy numbers in the fuzzy
literature\footnote{$T_{a,b,c,d}\left(  x\right)  $ is just a trapezoidal
function of parameters $\left(  a,b,c,d\right)  $ whilst $S_{\alpha,\gamma}$
is defined as:
\[
S_{\alpha,\gamma}\left(  x\right)  =\left\{
\begin{tabular}
[c]{ll}%
$0$ & $x<\alpha$\\
$2\left(  \frac{\left(  x-\alpha\right)  }{\left(  \gamma-\alpha\right)
}\right)  ^{2}$ & $\alpha<x\leq\frac{\alpha+\gamma}{2}$\\
$1-2\left(  \frac{\left(  x-\gamma\right)  }{\left(  \gamma-\alpha\right)
}\right)  ^{2}$ & $\frac{\alpha+\gamma}{2}<x\leq\gamma$\\
$1$ & $\gamma<x$%
\end{tabular}
\ \ \right.
\]
\par
{}}, which are depicted in Figure \ref{FiguQuantifiers}. For example, the
semi-fuzzy quantifier $\mathbf{about\_5}\left(  Y_{1},Y_{2}\right)  $ just
returns the evaluation of the cardinality of $Y_{1}\cap Y_{2}$ in the function
depicted in Figure \ref{FiguQuantifiers}, a).%

%TCIMACRO{\FRAME{ftbpFU}{5.3225cm}{1.9169cm}{0pt}{\Qcb{Fuzzy numbers for
%modelling \QTR{it}{`around 5'} and \QTR{it}{`at lest 80\%'}.}}%
%{\Qlb{FiguQuantifiers}}{alrededorormor80percent.eps}%
%width 5.3225cm;  height 1.9169cm;  depth 0pt;  original-width 8.4665in;
%original-height 2.9713in;  cropleft "0";  croptop "1";  cropright "1";
%filename 'Imagenes/AlrededorOrMor80percent.eps';file-properties "XNPEU";}} }%
%BeginExpansion
%EndExpansion

Semi-fuzzy quantifiers are easier to define and interpret than fuzzy
quantifiers, but they are not powerful enough to evaluate fuzzy quantified
expressions, as they only accept classic inputs. To solve that, in
\cite[definition 2.8]{Glockner06Libro} mechanisms to transform semi-fuzzy
quantifiers into fuzzy quantifiers were proposed (i.e., mappings with domain
in the universe of semi-fuzzy quantifiers and range in the universe of fuzzy quantifiers):

\begin{definition}
\cite{Glockner06Libro} A quantifier fuzzification mechanism (QFM)
$\mathcal{F}$ assigns to each semi-fuzzy quantifier $Q:\mathcal{P}\left(
E\right)  ^{n}\rightarrow\mathbf{I}$ a corresponding fuzzy quantifier
$\mathcal{F}\left(  Q\right)  :\widetilde{\mathcal{P}}\left(  E\right)
^{n}\rightarrow\mathbf{I}$ of the same arity $n\in\mathbb{N}$ and on the same
base set $E$.
\end{definition}

\section{Some probabilistic QFMs\label{Apendice}}

In this section we will present some \textit{QFMs }that can be interpreted
from a probabilistic point of view. In \cite{DiazHermida17-FuzzySets} a
thoroughly comparison of these models with other of the main QFMs that have
been presented in the literature can be consulted. The work \cite{Delgado14}
reviews the main approaches to fuzzy quantification that have been proposed,
comparing them against a list of criteria that do not include some of the
properties that have been used in \cite{DiazHermida17-FuzzySets}.

\subsection{Alpha-cut based QFMs\textbf{ }$\mathcal{F}^{I}$ and $\mathcal{F}%
^{MD}$}

In this section we will present two QFMs which are based on
\textit{alpha-cuts} (see below) of the input sets. Both of them admit a
probabilistic interpretation of fuzzy sets:

\begin{definition}
\cite{DiazHermida02-FuzzySets}, \cite{DiazHermida06Tesis} Let $Q:\mathcal{P}%
\left(  E\right)  ^{n}\rightarrow\mathbf{I}$ be a semi-fuzzy quantifier over a
base set $E$. The QFM $\mathcal{F}^{MD}$ is defined as:
\[
\mathcal{F}^{MD}\left(  Q\right)  \left(  X_{1},\ldots,X_{n}\right)  =\int
_{0}^{1}Q\left(  \left(  X_{1}\right)  _{\geq\alpha},\ldots,\left(
X_{n}\right)  _{\geq\alpha}\right)  d\alpha
\]
for every $X_{1},\ldots,X_{n}\in\widetilde{\mathcal{P}}\left(  E\right)  $,
where $\left(  X_{i}\right)  _{\geq\alpha}=\left\{  e\in E:\mu_{X_{i}}\left(
e\right)  \geq\alpha\right\}  $ is the alpha-cut of level $\alpha$ of $X_{i}$.
\end{definition}

For normalized fuzzy sets, $\mathcal{F}^{MD}$ coincides with the
quantification model $GD$ defined in \cite{Delgado99}, \cite{Delgado00} for
quantified expressions following the Zadeh's framework.

The definition of the $\mathcal{F}^{I}$ model is presented now:

\begin{definition}
\cite{DiazHermida00}, \cite{DiazHermida02-FuzzySets},\cite{DiazHermida06Tesis}
Let $Q:\mathcal{P}\left(  E\right)  ^{n}\rightarrow\mathbf{I}$ be a semi-fuzzy
quantifier over a base set $E$. The QFM $\mathcal{F}^{I}$ is defined as:
\begin{align*}
&  \mathcal{F}^{I}\left(  Q\right)  \left(  X_{1},\ldots,X_{n}\right) \\
&  =\int_{0}^{1}\ldots\int_{0}^{1}Q\left(  \left(  X_{1}\right)  _{\geq
\alpha_{1}},\ldots,\left(  X_{n}\right)  _{\geq\alpha_{n}}\right)  d\alpha
_{1}\ldots d\alpha_{n}%
\end{align*}
for every $X_{1},\ldots,X_{n}\in\widetilde{\mathcal{P}}\left(  E\right)  $.
\end{definition}

\subsection{QFM $\mathcal{F}^{A}$}

The QFM $\mathcal{F}^{A}$ fulfills the axiomatic framework presented in
\cite{Glockner06Libro}, although it does not belong to the class of\ `standard
DFSs' proposed by the author, being a `non-standard DFS'. This model can also
be interpreted in a probabilistic way, although it also accepts a definition
purely based on fuzzy operators, without reference to the probability theory
\cite{DiazHermida06Tesis}, \cite{DiazHermida17-FuzzySets},
\cite{DiazHermida10Arxiv}.

\begin{definition}
Let $X\in\widetilde{\mathcal{P}}\left(  E\right)  $ be a fuzzy set, $E$
finite. The probability of the crisp set $Y\in\mathcal{P}\left(  E\right)  $
of being a representative of the fuzzy set $X\in\widetilde{\mathcal{P}}\left(
E\right)  $ is defined as%
\[
m_{X}\left(  Y\right)  =%
%TCIMACRO{\dprod \limits_{e\in Y}}%
%BeginExpansion
{\displaystyle\prod\limits_{e\in Y}}
%EndExpansion
\mu_{X}\left(  e\right)
%TCIMACRO{\dprod \limits_{e\in E\backslash Y}}%
%BeginExpansion
{\displaystyle\prod\limits_{e\in E\backslash Y}}
%EndExpansion
\left(  1-\mu_{X}\left(  e\right)  \right)
\]

\end{definition}

Previous definition is used to define the $\mathcal{F}^{A}$ DFS:

\begin{definition}
\cite{DiazHermida04IPMU} Let $Q:\mathcal{P}\left(  E\right)  ^{n}%
\rightarrow\mathbf{I}$ be a semi-fuzzy quantifier, $E$ finite. The DFS
$\mathcal{F}^{A}$ is defined as%
\begin{align*}
&  \mathcal{F}^{A}\left(  Q\right)  \left(  X_{1},\ldots,X_{n}\right)  \\
&  =\sum_{Y_{1}\in\mathcal{P}\left(  E\right)  }\ldots\sum_{Y_{n}%
\in\mathcal{P}\left(  E\right)  }m_{X_{1}}\left(  Y_{1}\right)  \ldots
m_{X_{n}}\left(  Y_{n}\right)  Q\left(  Y_{1},\ldots,Y_{n}\right)
\end{align*}
for all $X_{1},\ldots,X_{n}\in\widetilde{\mathcal{P}}\left(  E\right)  $.
\end{definition}

\section{Fuzzy quantification to model quantified patterns in the temporal
domain}

This section deals with the use of fuzzy quantification in a `descriptive
sense', where the objective of fuzzy quantifiers is simply to check the degree
of fulfilment of a particular linguistic expression. The different examples we
will consider are aim to model quantified temporal expressions, as they
represent one of the best examples of the use of fuzzy quantifiers for
modelling the semantics of natural language. Before proceeding, we should take
into account that to express the semantics of quantified expressions, we will
only need to define the convenient semi-fuzzy quantifiers (which take as
inputs binary arguments), as we will rely on QFMs like the ones presented in
section \ref{Apendice} to convert these semi-fuzzy quantifiers into fully
operational fuzzy quantifiers.

\begin{figure}
[ptb]
\begin{center}
\includegraphics[
height=2.2543cm,
width=5.371cm
]%
{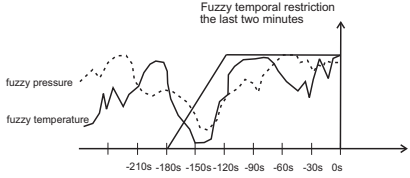}%
\caption{Example of two fuzzy signals and a temporal expression.}%
\label{FiguTemporalExpression}%
\end{center}
\end{figure}

\begin{itemize}
\item \textbf{Proportional temporal case, }which are useful to evaluate
expressions fitting the pattern \textit{\textquotedblleft}$Q$\textit{ }%
$E$\textit{ in }$T$\textit{ are }$Y$\textit{\textquotedblright\ }or the
pattern \textit{\textquotedblleft}$Q$\textit{ }$E$\textit{ in }$T$\textit{
fulfilling }$Y_{1}$ \textit{are }$Y_{2}$, where $Q$ is a proportional
semi-fuzzy quantifier, $T$ is a temporal reference and $Y,Y_{1},Y_{2}$ are
binary time series. An example of an expression fitting the first pattern is
\textquotedblleft\textit{most days in the last weeks were hot}", whilst
\textquotedblleft\textit{most hot days in the last weeks were associated to
high humidity values}\textquotedblright\ fits the second pattern. This
situation can be modeled by means of the following semi-fuzzy quantifiers:%
\begin{align*}
Q\left(  T,Y\right)   &  =\left\{
\begin{array}
[c]{cc}%
f_{Q}\left(  \frac{\left\vert T\cap Y\right\vert }{\left\vert T\right\vert
}\right)  & T\neq\varnothing\\
1 & T=\varnothing
\end{array}
\right. \\
Q\left(  T,Y_{1},Y_{2}\right)   &  =\left\{
\begin{array}
[c]{cc}%
fn\left(  \frac{\left\vert T\cap Y_{1}\cap Y_{2}\right\vert }{\left\vert T\cap
Y_{1}\right\vert }\right)  & T\cap Y_{1}\neq\varnothing\\
1 & T\cap Y_{1}=\varnothing
\end{array}
\right.
\end{align*}
where $f_{Q}$ is a proportional fuzzy number, like the one presented in Figure
\ref{FiguQuantifiers}, b). In Figure \ref{FiguTemporalExpression}, two fuzzy
signals and a fuzzy temporal reference are depicted. Semi-fuzzy quantifiers as
the ones previously defined, could be applied to these inputs after
transforming them into fuzzy quantifiers by means of a QFM.
\end{itemize}

%

%TCIMACRO{\FRAME{ftbpFU}{5.371cm}{2.2543cm}{0pt}{\Qcb{Example of two fuzzy
%signals and a temporal expression.}}{\Qlb{FiguTemporalExpression}%
%}{temporalexample.eps}{\special{ language "Scientific Word";  type "GRAPHIC";
%width 5.371cm;  height 2.2543cm;  depth 0pt;  original-width 3.6613in;
%original-height 1.5085in;  cropleft "0";  croptop "1";  cropright "1";
%filename 'Imagenes/TemporalExample.eps';file-properties "XNPEU";}} }%
%BeginExpansion
%EndExpansion

\begin{itemize}
\item \textbf{Similarity temporal case:} which are useful to evaluate
expressions fitting the pattern \textit{\textquotedblleft In }$T,$\textit{
}$Y_{1}$ \textit{and }$Y_{2}$ are $Q$ similar\textit{\textquotedblright.} An
example of an expression fitting this pattern is \textquotedblleft\textit{in
the last weeks, hot temperatures and high humidity values happened together
about the 80\% or more of the days}\textquotedblright. This situation can be
modelled by means of the following semi-fuzzy quantifier:%
\[
Q\left(  T,Y_{1},Y_{2}\right)  =\left\{
\begin{array}
[c]{cc}%
fn\left(  \frac{\left\vert T\cap Y_{1}\cap Y_{2}\right\vert }{\left\vert
T\cap\left(  Y_{1}\cup Y_{2}\right)  \right\vert }\right)  & T\cap\left(
Y_{1}\cup Y_{2}\right)  \neq\varnothing\\
1 & T\cap\left(  Y_{1}\cup Y_{2}\right)  =\varnothing
\end{array}
\right.
\]
where $f_{Q}$ is also a proportional fuzzy number.
\end{itemize}

More complex semi-fuzzy quantifiers could be defined for other situations.

In practical problems, it is common that we would like to check a quantified
temporal pattern for the whole temporal axis. To deal with this situation we
will introduce some notation that will allow us to define a quantified
temporal pattern relative to a moving temporal window.

Let $FT$ be a relative temporal fuzzy number defined with respect to a
temporal point $0$ (e.g., the temporal reference in Figure
\ref{FiguTemporalExpression}). The idea of $FT$ being relative is to work as a
reference fuzzy number that we can displace over one or several temporal
signals. On the basis of the fuzzy number $FT$, we define the temporal fuzzy
number $FT^{t_{0}}$ relative to the instant $t_{0}$ as:%

\[
FT^{t_{0}}\left(  t\right)  =FT\left(  t-t_{0}\right)
\]

Now, let us suppose $\widetilde{Q}:\widetilde{\mathcal{P}}\left(  E\right)
^{n}\longrightarrow\left[  0,1\right]  $ is a fuzzy quantifier like the ones
defined before, where we are supposing the first argument refers to the
temporal constraint. We define the application of the fuzzy quantified pattern
to the temporal axis as:%
\[
R_{S_{1},\ldots,S_{n}}^{\widetilde{Q},FT}\left(  t_{i}\right)  =\widetilde
{Q}\left(  FT^{t_{i}},S_{1},\ldots,S_{n}\right)
\]
where $FT^{t_{i}}$ is the displacement of $FT$ by $t_{i}$ towards the right,
and $S_{1},\ldots,S_{n}:T\longrightarrow\left[  0,1\right]  $ are fuzzy signals.

We also could suppose some kind of temporal displacement between the
application of the temporal constraint for the different input signals. For
example, for the proportional case with two signals and a temporal reference:%

\[
R_{S_{1},S2}^{\widetilde{Q},FT,d}\left(  t_{i}\right)  =\widetilde{Q}\left(
FT^{t_{i}},S_{1},S_{2}^{d}\right)
\]
where by $S_{2}^{d}\left(  t\right)  $ we are representing the displacement of
$S$ by $d$ temporal units.

\begin{figure}
[ptb]
\begin{center}
\includegraphics[
height=4.538cm,
width=5.5903cm
]%
{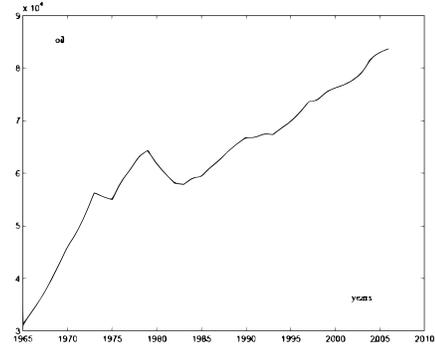}%
\caption{Daily world oil production in the period 1965-2006 (barrels)}%
\label{FiguOil}%
\end{center}
\end{figure}

Let us show now an example of the application of a fuzzy quantified expression
to a time series. In Figure \ref{FiguOil}, the daily world oil production in
the period 1965-2006 is represented. For this example, we will evaluate the
quantified expression \textquotedblleft\textit{in most of the last five years,
increments in oil production were negative or only slightly superior to
0}\textquotedblright, which could be modeled by means of the following expression:%

\begin{equation}
R_{ns\_oil}^{\widetilde{most},last\_five\_years}\left(  t\right)
=\widetilde{most}\left(  last\_five\_years^{t},nsoil\right)  \label{EqFQOil}%
\end{equation}
%

%TCIMACRO{\FRAME{ftbpFU}{5.5903cm}{4.538cm}{0pt}{\Qcb{Daily world oil
%production in the period 1965-2006 (barrels)}}{\Qlb{FiguOil}}%
%{1_oilproduction.eps}{\special{ language "Scientific Word";  type "GRAPHIC";
%width 5.5903cm;  height 4.538cm;  depth 0pt;  original-width 6.6102in;
%original-height 5.3499in;  cropleft "0";  croptop "1";  cropright "1";
%filename 'Imagenes/1_oilProduction.eps';file-properties "XNPEU";}} }%
%BeginExpansion
%EndExpansion

where \textit{most} is a binary proportional fuzzy quantifier defined as
$\mathcal{F}^{A}\left(  most\right)  $, being $most$ the semi-fuzzy
quantifier:%
\[
most\left(  T,S\right)  =\left\{
\begin{array}
[c]{cc}%
S_{0.7,0.9}\left(  \frac{\left\vert T\cap S\right\vert }{\left\vert
T\right\vert }\right)  & T\neq\varnothing\\
1 & T=\varnothing
\end{array}
\right.
\]

$last\_five\_years$ is a fuzzy set defining the temporal constraint:
$last\_five\_years\left(  x\right)  =T_{-8,-5,0,0}\left(  x\right)  $, and
$nsoi\not l \left(  t\right)  $ is the fuzzy signal that results of applying
the fuzzy number $S_{1,4}\left(  t\right)  $ to the percentage variations in
oil production:%

\[
nsoil\left(  t\right)  =S_{1,4}^{left}\left(  100\cdot\frac{oil\left(
i\right)  -oil\left(  i-1\right)  }{oil\left(  i-1\right)  }\right)
\]

\begin{figure}
[ptb]
\begin{center}
\includegraphics[
height=4.1964cm,
width=5.371cm
]%
{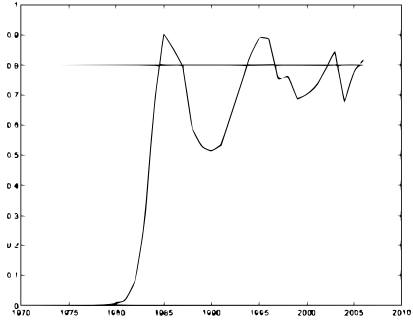}%
\caption{Evaluation of "\textit{in most years preceding }$t_{i}$\textit{,
increasing in oil production were negative or only slightly positive}".}%
\label{FiguEvaluationOil}%
\end{center}
\end{figure}

We show in Figure \ref{FiguEvaluationOil} the result of evaluating
\ref{EqFQOil} for the dataset in Figure \ref{FiguOil}. A threshold value 0.8
is depicted for indicating the years fulfilling the expression. Since, for
example, for the year 1995 the threshold is surpassed, we could intrepret that
\textquotedblleft\textit{in most of the years preceding 1995, increments in
oil production were negative or slightly positive\textquotedblright}%

%TCIMACRO{\FRAME{ftbpFU}{5.371cm}{4.1964cm}{0pt}{\Qcb{Evaluation of
%"\QTR{it}{in most years preceding }$t_{i}$\QTR{it}{, increasing in oil
%production were negative or only slightly positive}".}}%
%{\Qlb{FiguEvaluationOil}}{3_evaluationoil.eps}%
%width 5.371cm;  height 4.1964cm;  depth 0pt;  original-width 3.3466in;
%original-height 2.6044in;  cropleft "0";  croptop "1";  cropright "1";
%filename 'Imagenes/3_evaluationOil.eps';file-properties "XNPEU";}} }%
%BeginExpansion
%EndExpansion

In \cite{DiazHermida11FOCI}, the use of linguistic quantified patterns was
also presented with two other objectives: \textit{differentiation }and
\textit{aggregation}. Let us consider a set of elements $E=\left\{
e_{1},\ldots,e_{n}\right\}  \ $for which a specific temporal quantified
pattern can be applicable to some of the properties of the $e_{i}s$. For
example, $E$ could be a set of clients of an energy company and the quantified
pattern: \textit{\textquotedblleft in} \textit{nearly all the days of last
month, energy consumption in the morning was higher than consumption in the
afternoon for client }$e_{i}$\textquotedblright. By means of differentiation,
we try to detect elements that do not follow the common behavior. For example,
if most of the elements in $E$ fulfill the quantified pattern with a high
degree, then we will search for specific $e_{i}s$ with a low degree of
fulfillment. Symmetrically, we could look for $e_{i}s$ fulfilling the pattern
with a high degree of fulfillment when most elements do not fulfill the
pattern. This could be useful to identify anomalous patterns expressed as
linguistic quantified statements.

In the case of aggregation we also consider a series of elements $E$ and a
quantified pattern that can be evaluated for some property of the $e_{i}s$. In
this case, the objective is to summarize the general fulfillment of the
pattern by the elements of $E$ (i.e., \textit{\textquotedblleft for most
}$e_{i}s$\textit{ the pattern is fulfilled\textquotedblright}). For example,
to deal with linguistic expressions like \textit{\textquotedblleft almost all
the days of the last month, consumption in the morning was higher than
consumption in the afternoon for most clients of the company\textquotedblright%
}.

\section{Summarizing data with fuzzy quantifiers}

The objective of summarizing data with fuzzy quantifiers is to compute a
single quantifier or a set of quantified expressions that adequately summarize
a set of data. Let us suppose there is a set of data $E=\left\{  e_{1}%
,\ldots,e_{m}\right\}  $ (e.g., students) and a set of numerical attributes
$a_{1}\,,\ldots,a_{r}$ (e.g., `age', `height') that can be applied to the
elements in $E$. For each attribute $a_{j}$, we will assume that a linguistic
variable $L_{j}=\left\{  l_{j,1},\ldots,l_{j,p_{j}}\right\}  $ has been
defined to adapt the numerical attributes to linguistic values. Moreover, in
some cases we will also suppose there is a predefined fuzzy quantified
partition $F_{Q}=\left\{  f_{Q_{0}},\ldots,f_{Q_{W-1}}\right\}  $ of the
proportional universe $\left[  0,\ldots,100\%\right]  $ (e.g., `nearly none',
`a few',\ `several', `many', `nearly all').

Several options arise to build a summary of the input data based on fuzzy
quantification, depending on the structure of the summaries and the
consideration of a possible predefined quantified partition. For example, we
could be interested in building a summary composed of a unique quantified
expression, limiting us to summarize the fulfillment of an specific label
(e.g. \textit{\textquotedblleft most people are young\textquotedblright}) or
to explain the data by means of several quantified expressions dealing
simultaneously with several labels (e.g. \textit{\textquotedblleft some people
are young and some are old\textquotedblright}). In the following sections, we
will present some of the proposals that have been previously published to
handle this problem.

\subsection{Computing a unique quantifier to summarize the data}

We will consider two different options to compute a quantified label
summarizing a set of data.

\subsubsection{Computing the best quantifier within an existing quantified
partition}

Let us suppose first that there exist a predefined quantified partition
$F_{Q}$ in which we should base our summary and that we want to summarize the
data with respect to the fulfillment of a fixed set of linguistic labels
$l_{1},\ldots,l_{n}$, where label $l_{i}$ is applied to the attribute $a_{i}$.
For example, considering two properties `age' and `height' and a binary
proportional quantifier, $l_{1}$ could be `tall' applied to heightness and
$l_{2}$ could be\ `normal' applied to weigthness.

In this case, the most reasonable option is to return the quantified label
which provides the greater degree of fulfillment. For example, if for a given
set of students we obtained the higher degree of fulfillment for the fuzzy
quantifier `\textit{most' }for the fixed set of labels `tall' and `heavy', we
could summarize the data as \textit{\textquotedblleft most tall students are
heavy\textquotedblright}.

Summarizing a set of data by means of a unique quantified label can be
inadequate in the case there is not a unique quantified label with a high
degree of fulfillment, or if several quantified labels share a similar degree
of fulfillment. In this case, a convenient answer could be to indicate that
none of the quantified labels is adequate to summarize the data.

\subsubsection{Computing the optimal quantifier to summarize a set of data}

In this case, we will not constraint us to a set of predefined labels, being
the objective to compute automatically the quantified label that better
summarize the data. This problem was addressed in \cite{Glockner06SMPS} where
an algorithm solution was provided to compute the optimal crisp proportional
quantifier following the expression:%
\begin{equation}
rate_{\left[  r_{1},r_{2}\right]  }\left(  Y_{1},Y_{2}\right)  =\left\{
\begin{array}
[c]{ccc}%
1 & : & Y_{1}\neq\varnothing\wedge\frac{\left\vert Y_{1}\cap Y_{2}\right\vert
}{\left\vert Y_{1}\right\vert }\in\left[  r_{1},r_{2}\right]  \\
0 & : & else
\end{array}
\right.  \label{Glockner_rate}%
\end{equation}

The semantics of the previous quantifier is associated to expressions like
\textit{\textquotedblleft between }$r_{1}$ \textit{and }$r_{2}$
\textit{percent of the} $X_{1}$'$s$ \textit{are }$X_{2}$'$s$\textquotedblright%
. When $r_{2}=100\%$, the semantics is \textit{\textquotedblleft at least}
$r_{1}$ \textit{percent of the }$X_{1}$'$s$ \textit{are }$X_{2}$%
'$s$\textquotedblright.

The model proposed in \cite{Glockner06SMPS}, uses a predefined parameter
$\delta_{\max}$ to restrict the amplitude of the semi-fuzzy proportional
quantifier (i.e. $\delta_{\max}=0.2$ limits $r_{2}-r_{1}$ to $0.2$, or in
proportional terms a percentage range equal or inferior to $20\%$).
Constrained by this parameter it computes the optimal (in the sense of
producing the highest evaluation degree) semi-fuzzy quantifier following
expression \ref{Glockner_rate}. This algorithm was developed for the
$\mathcal{M}_{CX}$ DFS proposed by the author, although it is also valid for
the class of `standard DFSs' \cite{Glockner06Libro}. We cannot present the
details of the algorithm for lack of space as it depends heavily on the
properties of the $\mathcal{M}_{CX}$ DFS. Basically, given two fuzzy sets
$X_{1}$ and $X_{2}$, and a supposed parameter $\delta_{\max}=0.2$, the
proposal in \cite{Glockner06SMPS} permits the computation of expressions like
\textit{\textquotedblleft between 62.5\% and 75\% of the }$X_{1}$'$s$\textit{
are }$X_{2}$'$s$\textquotedblright, where `between 62.5\% and 75\%' is the
$rate$ quantifier which produces the highest evaluation degree for this
$\delta_{\max}$.

The author has not extended this proposal to other kinds of semi-fuzzy
quantifiers. However, the same ideas presented in the previous reference can
be used to adapt the algorithm to other kinds of semi-fuzzy quantifiers which
will allow us to search for other relationships between the data (e.g.
comparative quantifiers, etc.).

Although a similar proposal have not been presented for the probabilistic
models in section \ref{Apendice}, it is possible to approximate the optimal
$rate_{\left[  r_{1},r_{2}\right]  }$ quantifier for a given amplitude
$\delta_{\max}$ evaluating a series of rate quantifiers starting in
$rate_{\left[  0,\delta_{\max}\right]  }$, and displacing this semi-fuzzy
quantifier over the proportional axis following the pattern $rate_{\left[
h,\delta_{\max}+h\right]  }$. The summary will be constructed using the
quantifier for which the greatest evaluation value was obtained.

We have sketched some ideas for summarizing data by means of crisp
proportional quantifiers. However, the adaptation of previous ideas to learn
non crisp quantifiers, have not been dealt with in the literature to our knowledge.

\subsection{Computing a set of compatible quantified expressions to summarize
the data}

In \cite{DiazHermida10Estylf} a method was proposed to summarize a set of
temporal data by means of several compatible quantified expressions using the
$\mathcal{F}^{A}$ DFS. Let us consider the existence of a linguistic variable
(like the one in Figure \ref{FiguTemperatureLabels}) and a Ruspini unary
proportional quantified partition of the quantification universe\footnote{The
linguistic variable represented in figure \ref{FiguTemperatureLabels} is a
Ruspini partition as the membership degrees of the different labels adds to
$1$ for each point in the $x$ axis. A Ruspini quantified partition follows a
similar pattern for fuzzy quantifiers.}. In their proposal, the authors
computed the evaluation results of each possible pair of label/quantifier.
Pairs of label/quantifiers with a high evaluation result, should be good
candidates to summarize some characteristic of the data. Let us suppose a
temperature data set and that the computation of $\mathcal{F}^{A}\left(
Q_{i}\right)  \left(  l_{j}\left(  temperatures\right)  \right)  $ for each
possible $\left(  i,j\right)  $, with \textit{\textquotedblleft nn=nearly
none, f=a few, s=several, m=many }and\textit{ na=nearly all\textquotedblright}
produces the following evaluation matrix:

\begin{center}
$%
\begin{tabular}
[c]{|l|l|l|l|l|l|}\hline
{\tiny Months} & \multicolumn{5}{|l|}{{\tiny April}}\\\hline
{\tiny Labels%
%TCIMACRO{\TEXTsymbol{\backslash}}%
%BeginExpansion
$\backslash$%
%EndExpansion
Quantifiers} & {\tiny nn} & {\tiny f} & {\tiny s} & {\tiny m} & {\tiny na}%
\\\hline
{\tiny very low} & {\tiny 1} & {\tiny 0} & {\tiny 0} & {\tiny 0} &
{\tiny 0}\\\hline
{\tiny low} & {\tiny 0} & {\tiny 0.72} & {\tiny 0.28} & {\tiny 0} &
{\tiny 0}\\\hline
{\tiny warm} & {\tiny 0} & {\tiny 0} & {\tiny 0.28} & {\tiny 0.72} &
{\tiny 0}\\\hline
{\tiny hot} & {\tiny 1} & {\tiny 0} & {\tiny 0} & {\tiny 0} & {\tiny 0}%
\\\hline
{\tiny very hot} & {\tiny 1} & {\tiny 0} & {\tiny 0} & {\tiny 0} &
{\tiny 0}\\\hline
\end{tabular}
$
\end{center}

In their approach, the authors presented a greedy algorithm to extract a set
of quantified expressions taking as input the evaluation matrix and some
heuristics inspired in conversational intuitions. The possibility of merging
consecutive quantifiers has also been include in the proposal in
\cite{DiazHermida10Estylf}.

For the previous example, a possible summary could be
\textit{\textquotedblleft many temperatures were warm\textquotedblright} and
\textit{\textquotedblleft a few were low\textquotedblright. }As in the
previous case, more study is needed in order to advance in the methods for
summarizing the data by means of a set of compatible quantified expressions.

\section{Fuzzy quantification for machine learning}

In this section we will present two specific examples which prove the utility
of fuzzy quantifiers in machine learning applications.

\subsection{Fuzzy quantification for data mining of temporal constraint
networks}

In this section we will make a proposal to integrate fuzzy quantification into
temporal constraint data mining. This proposal is hypothetical, and it has not
been implemented yet.

Temporal constraint networks are temporal structures whose aim is to represent
the temporal occurrence of events constrained by some temporal metric between
them. Temporal constraint networks are represented by means of graphs, where
nodes represent the occurrence of events whilst arcs represent temporal
distances between nodes.

We will follow \cite{alvarezmr2013discovering} to introduce the idea. In this
reference, a specific proposal to mine fuzzy constraint networks inspired in
the Apriori algorithm for detecting association rules was proposed. The mining
process operates over a set of observables $O=\left\{  o_{1},\ldots
,o_{n}\right\}  $, or entities of the domain for which there exists an
observation procedure. Observation procedures identify the presence of an
observable in a temporal point. This abstract definition will allow us to
introduce fuzzy quantifiers as observables, in order to propose an idea to
mine fuzzy constraint networks were temporal entities are modelled by means of
fuzzy quantifiers.

In previous proposal two types of temporal entities are considered:

\begin{itemize}
\item An \textbf{event} is a tuple $\left(  O_{i},a=v,t\right)  $ where
$O_{i}\in\mathbf{O}$ is an observable, $a$ is an attribute with value $v\in
V\left(  a\right)  $ and $t\in\mathbb{N}$ is a time instant.

\item An \textbf{episode} is a tuple $\left(  O_{i},a=v,t_{b},t_{c}\right)  $
where $O_{i}\in\mathbf{O}$ is an observable, $a$ is an attribute with value
$v\in V\left(  a\right)  $ and $t_{b},t_{c}\in\mathbb{N}$ denote respectively
the begin and the end of the episode. In practice, episodes can be represented
by an initial and an ending event.
\end{itemize}

As we introduced before, fuzzy quantification can be introduced into temporal
data mining playing the role of \textit{observation procedures}. Let
$\mathbf{TP=}\left\{  tp_{1},\ldots,tp_{n}\right\}  $ be a finite set of
temporal patterns defined over our set of input signals $S_{1}\left(
t\right)  ,\ldots,S_{G}\left(  t\right)  \in\mathbf{S}$. For example, a
temporal pattern $tp_{k}$ could follow the scheme:%
\[
tp_{k}\left(  t_{i}\right)  =\widetilde{Q}\left(  FT^{t_{i}},S_{1}\right)
\]
for a unary quantifier, or%

\[
tp_{k}\left(  t_{i}\right)  =\widetilde{Q}\left(  FT^{t_{i}},S_{1},S_{2}%
^{d}\right)
\]
for a binary proportional quantifier with displacement $d$. These patterns
could be predefined by the expert guiding the mining process or generated by
means of an automatic procedure.

Constraining the fuzzy patterns by means of some threshold (e.g., assuming the
pattern is fulfilled in $t$ if its degree of fulfillment is superior to $0.8$,
and that is not fulfilled in other case) we can introduce them in constraint
data mining algorithms as binary observables. For example, a temporal pattern
could be \textit{\textquotedblleft the temperature was extremely high in the
last five minutes\textquotedblright\ }or \textit{\textquotedblleft most high
pressure values in the last half an hour were associated to extremely high
temperature values}\textquotedblright.

\begin{figure}
[ptb]
\begin{center}
\includegraphics[
height=3.2222cm,
width=6.8197cm
]%
{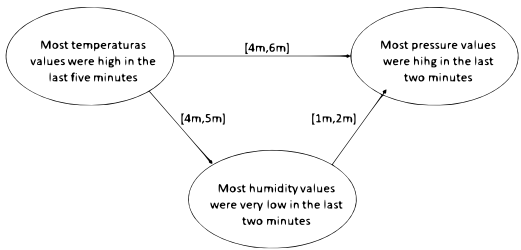}%
\caption{Example of a hypothetical constraint network between quantified
temporal expressions.}%
\label{FiguNetwork}%
\end{center}
\end{figure}

In Figure \ref{FiguNetwork} we show a hypothetical example of a possible
quantified constraint temporal network that could be obtained with this kind
of approximation. As future work, we will analyze the interest of this proposal.%

%TCIMACRO{\FRAME{ftbpFU}{6.8197cm}{3.2222cm}{0pt}{\Qcb{Example of a
%hypothetical constraint network between quantified temporal expressions.}%
%}{\Qlb{FiguNetwork}}{network.eps}{\special{ language "Scientific Word";
%type "GRAPHIC";  maintain-aspect-ratio TRUE;  display "USEDEF";
%valid_file "F";  width 6.8197cm;  height 3.2222cm;  depth 0pt;
%original-width 12.2972in;  original-height 5.7434in;  cropleft "0";
%filename 'Imagenes/network.eps';file-properties "XNPEU";}} }%
%BeginExpansion
%EndExpansion

\subsection{Systems of quantified fuzzy rules for classification and
regression}

In \cite{ismaelrodriguezfdez2011iterative} the learning of fuzzy controllers
in mobile robotics by means of quantified fuzzy rules was proposed. In fuzzy
control, a fuzzy controller is composed of a set of rules fulfilling the pattern:

\begin{center}
{\small (R1) If }$x_{1}$ {\small is }$l_{1,1}${\small and }$x_{2}$ {\small is
}$l_{1,2}$ {\small and ... and }$x_{n}$ {\small is }$l_{1,n}$ {\small then
}$y$ {\small is }$O_{1}$

{\small (R2) If }$x_{1}$ {\small is }$l_{2,1}${\small and }$x_{2}$ {\small is
}$l_{2,2}$ {\small and ... and }$x_{n}$ {\small is }$l_{2,n}$ {\small then
}$y$ {\small is }$O_{2}$

{\small ...}

{\small (Rm) If }$x_{1}$ {\small is }$l_{m,1}${\small and }$x_{2}$ {\small is
}$l_{m,2}$ {\small and ... and }$x_{n}$ {\small is }$l_{m,n}$ {\small then
}$y$ {\small is }$O_{m}$
\end{center}

where $x_{i}$ are inputs (e.g. signal values), $y$ is the output and $l_{i,j}$
and $O_{i}$ are linguistic labels. An example of a possible fuzzy rule could
be \textit{\textquotedblleft If the temperature is low and the pressure is
high then the velocity should be high\textquotedblright. }The idea of fuzzy
control is that if we observe the input values \textquotedblleft$x_{1}$ is
$P_{1}^{\ast}$ and $x_{2}$ is $P_{2}^{\ast}$ and ... and $x_{n}$ is
$P_{n}^{\ast}$\textquotedblright\ that do not fit exactly any of the rules of
the system, but we can guarantee a partial match with, let us suppose, the
rule $R_{i}$, we still can make some kind of affirmation about the output
based on the partial fulfillment. In fuzzy control systems, several rules can
be partially active at the same time. Different aggregation procedures are
available to integrate the output of the rules of the system and retrieve an
specific output value. Moreover, polynomial outputs (Takagi-Sugeno systems)
are a relevant variant of fuzzy controllers.

Fuzzy quantifiers can be used in fuzzy rule systems to introduce a new
aggregation level. In previous example, if we were considering fuzzy signals,
each atom (e.g., $x_{i}$ is $l_{i,j}$) would be applied to an specific
instant. But by means of fuzzy quantifiers we can substitute simple atoms by
quantified ones, allowing expressions like \textit{\textquotedblleft most
temperatures were high in the last minutes\textquotedblright}.

This approach was followed in \cite{ismaelrodriguezfdez2011iterative} for the
automatic learning of fuzzy controllers in mobile robotics. The idea of the
solution proposed in the author's approach, was to use fuzzy quantifiers as a
mean to aggregate `low level input variables' (variables with a small single
contribution to the system, as the distance of several laser beams). Given the
complexity of learning a complete set of rules involving quantifiers, a
genetic approach was proposed in which each individual codified a single rule.
The general idea of the author's approach was to learn the rule system rule by
rule, incorporating new rules based on different criteria.

The possibility of learning fuzzy rule system in fuzzy control proves the
capacity of fuzzy quantifiers to be integrated in fuzzy rule systems in
regression and classification problems. Learning fuzzy control systems is an
example of a regression procedure, in which input values are used to predict
an output value. As we are dividing the output axis by means of a linguistic
variable, classification can be associated to the selection of a specific
fuzzy label (e.g., the one that better includes the output value of the fuzzy
rule system).

\section{Conclusion}

In this paper we presented some of the different roles that fuzzy
quantification can play in data analytics and data mining. After introducing
the field of fuzzy quantification, we showed some uses of fuzzy quantifiers in
a `descriptive sense', with focus in the modelling of temporal expressions. We
continued presenting the application of fuzzy quantifiers for summarizing sets
of data by means of linguistic quantified expressions. Finally, two
applications of fuzzy quantifiers in machine learning, specifically in
temporal constraint networks and fuzzy systems of rules, were presented.

\section*{Aacknowledgement}
This work has received financial support from the Conseller\'{\i}a de Cultura,
Educaci\'{o}n e Ordenaci\'{o}n Universitaria (accreditation 2016-2019,
ED431G/08 and reference competitive group 2014-2017, GRC2014/030) and the
European Regional Development Fund (ERDF). Additionally, it was supported by
the Spanish Ministry of Economy and Competitiveness under the project TIN2015-73566-JIN.

\bibliographystyle{IEEEtran}

\begin{thebibliography}{10}
\providecommand{\url}[1]{#1}
\csname url@samestyle\endcsname
\providecommand{\newblock}{\relax}
\providecommand{\bibinfo}[2]{#2}
\providecommand{\BIBentrySTDinterwordspacing}{\spaceskip=0pt\relax}
\providecommand{\BIBentryALTinterwordstretchfactor}{4}
\providecommand{\BIBentryALTinterwordspacing}{\spaceskip=\fontdimen2\font plus
\BIBentryALTinterwordstretchfactor\fontdimen3\font minus
  \fontdimen4\font\relax}
\providecommand{\BIBforeignlanguage}[2]{{%
\expandafter\ifx\csname l@#1\endcsname\relax
\typeout{** WARNING: IEEEtran.bst: No hyphenation pattern has been}%
\typeout{** loaded for the language `#1'. Using the pattern for}%
\typeout{** the default language instead.}%
\else
\language=\csname l@#1\endcsname
\fi
#2}}
\providecommand{\BIBdecl}{\relax}
\BIBdecl

\bibitem{Zadeh65}
L.~Zadeh, ``Fuzzy sets,'' \emph{Information Control}, vol.~8, pp. 338--353,
  1965.

\bibitem{Zadeh75}
------, ``The concept of a linguistic variable and its application to
  approximate reasoning,'' \emph{Information Sciences}, vol.~1, pp. 119--249,
  1975.

\bibitem{Zadeh83}
------, ``A computational approach to fuzzy quantifiers in natural languages,''
  \emph{Comp. and Machs. with Appls.}, vol.~8, pp. 149--184, 1983.

\bibitem{Glockner06Libro}
I.~Gl{\"{o}}ckner, \emph{Fuzzy Quantifiers: A Computational Theory}.\hskip 1em
  plus 0.5em minus 0.4em\relax Springer, 2006.

\bibitem{Barwise81}
J.~Barwise and R.~Cooper, ``Generalized quantifiers and natural language,''
  \emph{Linguistics and Philosophy}, vol.~4, pp. 159--219, 1981.

\bibitem{Sanchez16}
M.~Ruiz, D.~S{\'{a}}nchez, and M.~Delgado, ``On the relation between fuzzy and
  generalized quantifiers,'' \emph{Fuzzy Sets and Systems}, p. In press, 2016.

\bibitem{DiazHermida17-FuzzySets}
F.~Diaz-Hermida, M.~Pereira-Fariña, J.~Vidal, and A.~Ramos-Soto,
  ``Characterizing quantifier fuzzification mechanisms: A behavioral guide for
  applications,'' \emph{Fuzzy Sets and Systems}, p. In press, 2017.

\bibitem{Delgado14}
M.~Delgado, D.~S{\'{a}}nchez, and M.~Vila, ``Fuzzy quantification: a state of
  the art,'' \emph{Fuzzy Sets and Systems}, vol. 242, pp. 1--302, 2014.

\bibitem{DiazHermida02-FuzzySets}
F.~D{\'{i}}az-Hermida, A.~Bugar{\'{i}}n, P.~Cari{\~{n}}ena, and S.~Barro,
  ``Voting model based evaluation of fuzzy quantified sentences: a general
  framework.'' \emph{Fuzzy Sets and Systems}, vol. 146, pp. 97--120, 2004.

\bibitem{DiazHermida06Tesis}
F.~D{\'{i}}az-Hermida, ``Modelos de cuantificaci\'on borrosa basados en una
  interpretaci\'on probabil\'istica y su aplicaci\'on en recuperaci\'on de
  informaci\'on,'' Ph.D. dissertation, Universidad de Santiago de Compostela,
  2006.

\bibitem{Delgado99}
M.~Delgado, D.~S\'anchez, and M.~A. Vila, ``A survey of methods for evaluating
  quantified sentences,'' in \emph{Proc. First European Society for fuzzy logic
  and technologies conference (EUSFLAT'99)}, 1999, pp. 279--282.

\bibitem{Delgado00}
------, ``Fuzzy cardinality based evaluation of quantified sentences,''
  \emph{International Journal of Approximate Reasoning}, vol.~23, no.~1, pp.
  23--66, 2000.

\bibitem{DiazHermida00}
F.~D{\'{i}}az-Hermida, A.~Bugar{\'{i}}n, P.~Cari{\~{n}}ena, and S.~Barro,
  ``Evaluaci\'on probabil{\'{i}}stica de proposiciones cuantificadas
  borrosas,'' in \emph{Actas del X Congreso Espa{\~{n}}ol Sobre
  Tecnolog{\'{i}}as y L\'ogica Fuzzy (ESTYLF 2000)}, 2000, pp. 477--482.

\bibitem{DiazHermida10Arxiv}
F.~Diaz-Hermida, A.~Bugar{\'{i}}n, and D.~E. Losada, ``The probatilistic
  quantifier fuzzification mechanism {FA}: {A} theoretical analysis,'' Centro
  Singular de Investigaci{\'{o}}n en Tecnolox{\'{i}}as da Informaci{\'{o}}n,
  Universidade de Santiago de Compostela, Tech. Rep., 2014, arXiv preprint
  arXiv: 1410.7233.

\bibitem{DiazHermida04IPMU}
F.~D{\'{i}}az-Hermida, D.~Losada, A.~Bugar{\'{i}}n, and S.~Barro, ``A novel
  probabilistic quantifier fuzzification mechanism for information retrieval,''
  in \emph{Proc. IPMU 2004, the 10th International Conference on Information
  Prosessing and Management of Uncertainty in Knowledge-Based Systems},
  Perugia, Italy, July 2004, pp. 1357,1364.

\bibitem{DiazHermida11FOCI}
F.~D{\'{i}}az-Hermida and A.~Bugar{\'{i}}n, ``Semi-fuzzy quantifiers as a tool
  for building linguistic summaries of data patterns,'' in \emph{In:
  Proceedings of IEEE SSCI2011 - 2011 IEEE Symposium of Computational
  Intelligence}, 2011, pp. 45--52.

\bibitem{Glockner06SMPS}
I.~Gl{\"{o}}ckner, ``Optimal selection of proportional bounding quantifiers in
  linguistic data summarization,'' in \emph{Proceedings of SMPS2006 - Special
  Session on Applications and Modelling of Imprecise Operators, Published in
  Soft Methods for Integrated Uncertainty Modelling, LNCS, Springer}, 2006, pp.
  173--181.

\bibitem{DiazHermida10Estylf}
F.~D\'{i}az-Hermida and A.~Bugar\'{i}n, ``Linguistic summarization of data with
  probabilistic fuzzy quantifiers,'' in \emph{XV Congreso Espa\~{n}ol Sobre
  Tecnolog\'{i}as y L\'{o}gica Fuzzy}, 2010, pp. 255--260.

\bibitem{alvarezmr2013discovering}
\BIBentryALTinterwordspacing
\'{A}lvarez MR, F.~P, and C.~{n}ena P, ``Discovering metric temporal constraint
  networks on temporal databases,'' \emph{Artificial Intelligence in Medicine},
  vol.~58, no.~3, pp. 139--154, 2013. [Online]. Available:
  \url{http://dx.doi.org/10.1016/j.artmed.2013.03.006}
\BIBentrySTDinterwordspacing

\bibitem{ismaelrodriguezfdez2011iterative}
\BIBentryALTinterwordspacing
I.~Rodr\'{i}guez-Fdez, M.~Mucientes, and A.~Bugar\'{i}n, ``Iterative rule
  learning of quantified fuzzy rules for control in mobile robotics,'' in
  \emph{5th IEEE International Workshop on Genetic and Evolutionary Fuzzy
  Systems}, 2011, pp. 111--118. [Online]. Available:
  \url{http://dx.doi.org/10.1109/GEFS.2011.5949500}
\BIBentrySTDinterwordspacing

\end{thebibliography}

\end{document}